%% file: main.tex
\documentclass[10pt,twocolumn,letterpaper]{article}

\usepackage[pagenumbers]{cvpr} 
\usepackage[accsupp]{axessibility}
\input{preamble}

\definecolor{cvprblue}{rgb}{0.21,0.49,0.74}
\usepackage[pagebackref,breaklinks,colorlinks,citecolor=cvprblue]{hyperref}
\usepackage[ruled, lined, linesnumbered, commentsnumbered, longend]{algorithm2e}
\SetAlFnt{\small}
\SetAlCapFnt{\small}
\SetAlCapNameFnt{\small}
\SetCommentSty{\small}
\SetAlgoNlRelativeSize{-1}
\SetAlgoNoEnd

\usepackage{algorithmic}
\algsetup{linenosize=\tiny}

\newcommand{\authorsep}{\hspace{8pt}}
\newcommand{\affiliationsep}{\hspace{8pt}}

\def\paperID{50}
\def\confName{CVPR}
\def\confYear{2024}

\title{Dr-SAM: An End-to-End Framework for Vascular Segmentation, Diameter Estimation, and Anomaly Detection on Angiography Images.}

\author{
Vazgen Zohranyan$^{1,4 \; *}$
\authorsep 
Vagner Navasardyan$^{2,6 \;\dag * }$
\authorsep
Hayk Navasardyan$^{3 \; *}$
\\
Jan Borggrefe$^{2,6 \; \dag}$
\authorsep
Shant Navasardyan$^{1,5}$
\\
{\small ${}^1$ Yerevan State University (YSU)
\affiliationsep
${}^2$ Ruhr-Universität Bochum (RUB)
\affiliationsep
${}^3$ Synopsys Armenia CJSC}
\\
{\small ${}^4$ ServiceTitan, Inc.
\affiliationsep
${}^5$ Picsart AI Research (PAIR)
\affiliationsep
${}^6$ Johannes Wesling University Hospital
}
\\
{\small \textbf{\url{https://github.com/vazgenzohranyan/Dr.SAM}}}
}

\date{\rule{0.9\linewidth}{0.2\linewidth}}
\begin{document}
\twocolumn[{%
\renewcommand\twocolumn[1][]{#1}%
\maketitle
\vspace{-10mm}
\begin{center}
    \centering
    \captionsetup{type=figure*}
    \includegraphics[width=\linewidth, height=9.45cm]{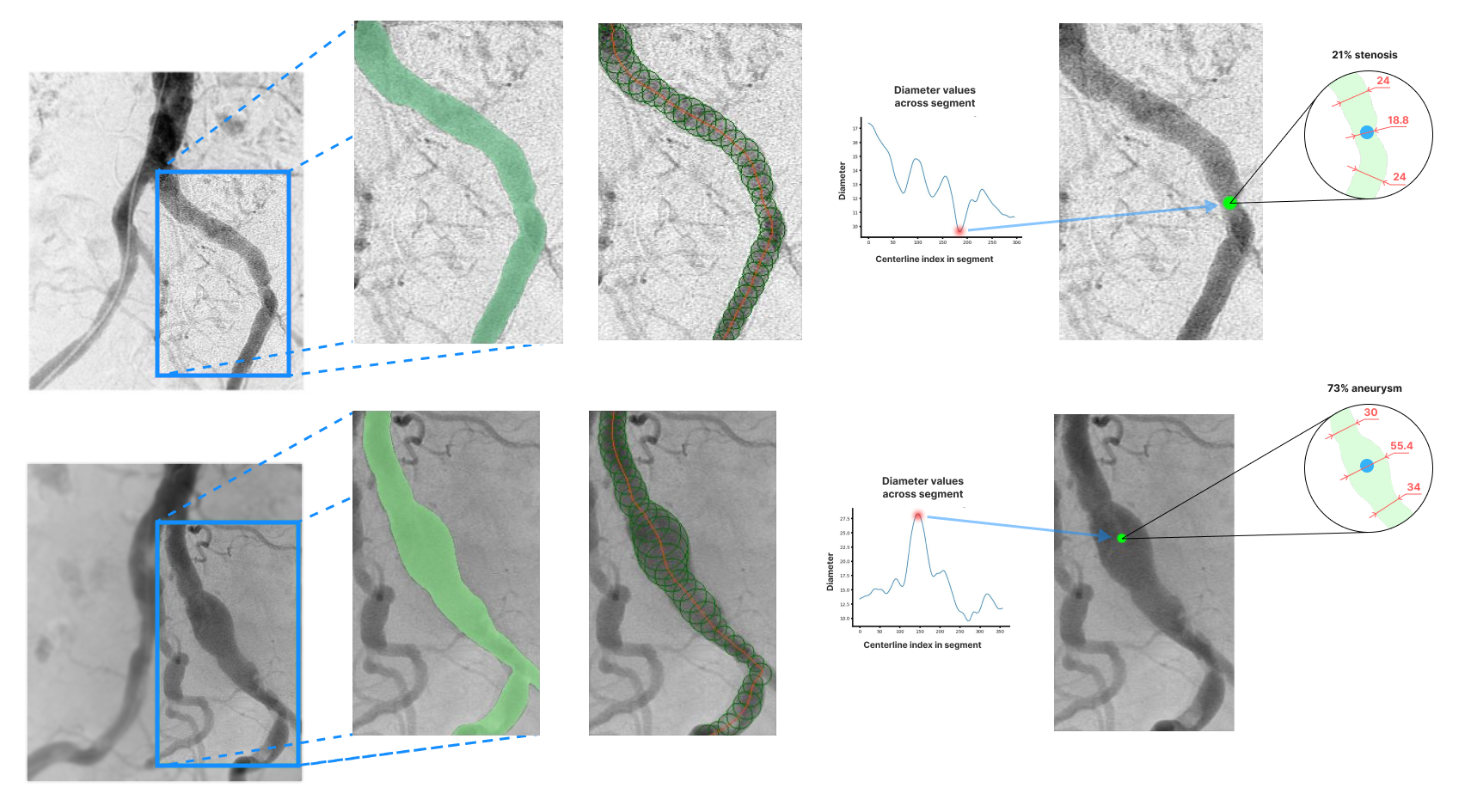}
    \vspace{-7mm} 
    \captionof{figure}[]{As an end-to-end framework for angiography image analysis, \textbf{\textit{Dr-SAM}} first extracts the segments of blood vessels, then detects the centerlines and estimates the diameters of the vessels (shown with circles), then, finally, recognizes the anomaly points indicating \textbf{\textit{stenoses}} or \textbf{\textit{aneurysms}} (green points).}
\end{center}%
}]
\def\thefootnote{*}\footnotetext{Equal contribution.}
\def\thefootnote{\dag}\footnotetext{Department of Radiology, Neuroradiology and Nuclear Medicine}

\input{sec/0_abstract}    
\input{sec/1_intro}

\input{sec/2_related_work}
\input{sec/3_method}
\input{sec/4_experiments}
\input{sec/5_conclusion}

{
    \small
    \bibliographystyle{ieeenat_fullname}
    \bibliography{main}
}

\end{document}

%% file: preamble.tex
\usepackage[dvipsnames]{xcolor}


%% file: sec/0_abstract.tex
\begin{abstract}
Recent advancements in AI have significantly transformed medical imaging, particularly in angiography, by enhancing diagnostic precision and patient care.
However existing works are limited in analyzing the aorta and iliac arteries, above all for vascular anomaly detection and characterization.
To close this gap, we propose Dr-SAM, a comprehensive multi-stage framework for vessel segmentation, diameter estimation, and anomaly analysis aiming to examine the peripheral vessels through angiography images.
For segmentation we introduce a customized positive/negative point selection mechanism applied on top of the Segment Anything Model (SAM), specifically for medical (Angiography) images. Then we propose a morphological approach to determine the vessel diameters followed by our histogram-driven anomaly detection approach.
Moreover, we introduce a new benchmark dataset for the comprehensive analysis of peripheral vessel angiography images which we hope can boost the upcoming research in this direction leading to enhanced diagnostic precision and ultimately better health outcomes for individuals facing vascular issues.
\end{abstract}

%% file: sec/1_intro.tex
\section{Introduction}

The blood supply to the lower body, including the legs and pelvic organs, relies heavily on the infrarenal aorta and pelvic arteries. Any narrowing (stenosis) \cite{Thiriet2015} or widening (aneurysms) in these vessels can lead to serious health issues. Angiography, an imaging technique that uses X-rays and contrast agents, is utilized for the precise diagnosis and treatment of these conditions. This imaging technique is particularly effective in identifying stenosis and aneurysms in the infrarenal aorta and pelvic arteries.
With the advancement of technology and the introduction of minimally invasive procedures, angiography has significantly enhanced the outcomes for patients with vascular diseases.
With the raise of AI angiography images got a chance to be analyzed semantically and assist the doctors more effectively in diagnosis forecasting.

 To conduct an angiographic examination, the doctor inserts a catheter into the arteries and through the catheter injects a contrast agent containing iodine into the blood vessel. The vessels can now be visualized using x-rays, usually in a substraction technique, to identify potential narrowing or widening. These images are used to evaluate the vessel diameter, stenoses or aneurysms, as well as the precise localization. If a relevant stenosis is detected during angiography, immediate treatment may be required, especially if it significantly impairs blood flow. In such cases, balloon or stent angioplasty may be an effective intervention. In this procedure, a small balloon at the end of the catheter is introduced to the narrowed area and then inflated to widen the narrowing and restore normal blood flow. For this reason, rapid and precise assessment of vascular diameters and their changes are crucial for stenosis/aneurysm detection and characterization.

 Simultaneous treatment of stenoses during angiographic examination offers a number of advantages. First, it can reduce the risk of complications that could arise if the patient had to return later for a separate operation. Second, it allows blood flow to be restored more quickly, minimizing the risk of tissue damage and complications such as tissue loss or necrosis. In addition, prompt treatment of stenosis may reduce the need for repeat interventions and improve long-term prognosis. For this reason, rapid and precise assessment of vascular changes is crucial.

 With the assistance of our tool, doctors can analyze images more quickly than manual examination allows. This efficiency shortens the time between diagnosis and the start of treatment, which is essential for conditions that need quick action. It also importantly minimizes the risk of diagnostic errors that can occur due to human factors like fatigue or subjective interpretation.

For this purpose we develop Dr-SAM, an end-to-end framework designed for vascular angiograpgy image analysis with vessel segmentation, diameter determination, and anomaly detection/characterization.

Various filter-based, learning-based, or regionally growing approaches have been developed for angiographic segmentation \cite{luo2023digital,digitalprocessing} including a wide usage of convolutional neural networks (CNNs \cite{deeplabv3plus2018,Unet}). 
CNNs have proven effective in segmentation across various applications and offer a potential solution to address the shortcomings of traditional methods in this complex area. 

\begin{figure}[t]
  \centering
   \includegraphics[width=0.8\linewidth]{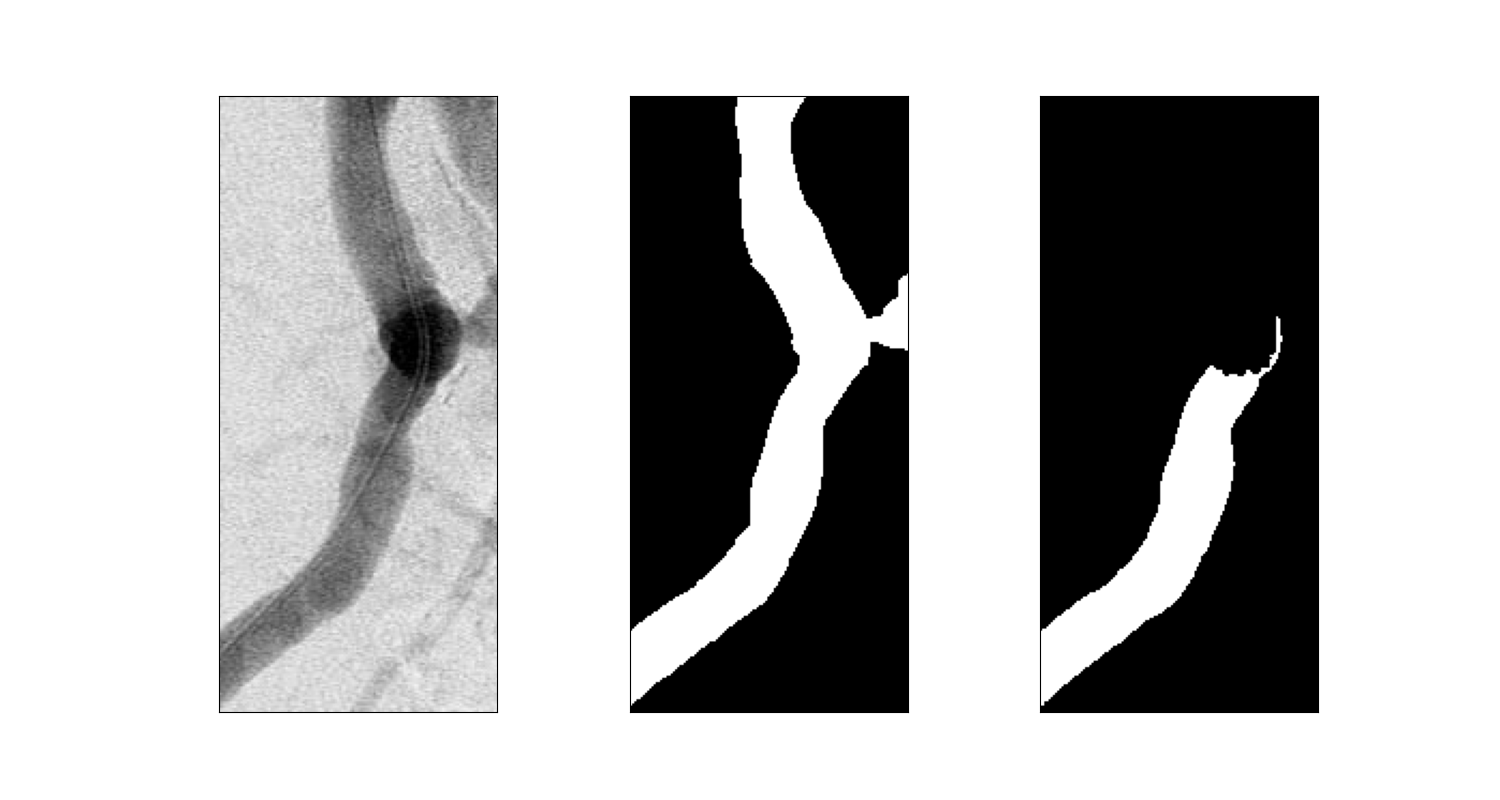}

   \caption{SAM result on X-Ray image without any prompts. Middle - ground truth mask, right - SAM predicted mask.}
   \label{fig:samoriginal}
\end{figure}

Recently, with the advancements of CNNs in the general domain segmentation task the Segment Anything Model (SAM)  \cite{SAM} was developed as an interactive tool for ultimate segmentation. However directly using SAM for vessel segmentation in angiography images usually leads to incorrect region selections (see Fig. \ref{fig:samoriginal}) due to the limitation of SAM requiring positive label points for precise segmentation. Therefore we designed a special positive point selection mechanism, tailored to use with SAM for the vascular angiography images.
 
 After segmenting the vessels in their corresponding regions, we further estimate vessel diameters and analyze stenosis/aneurysm anomalies. To achieve this, we utilize the topological skeleton of the binary mask by pruning certain branches. Due to the noise from the binary mask, the topological skeleton \cite{topoligicalskelton} may contain branches that are not actual vessel branches. Our algorithm identifies these branches by their size and prunes them, resulting in a clean tree-based vessel structure. This process improves the accuracy of approximate diameter estimation for vessel segments, aiding in the identification of stenosis/aneurysms.
 
 Furthermore, we introduce a benchmark dataset for the segmentation and anomaly detection on vascular angiography images crafted by domain specialists. We validate our approach on the proposed benchmark, and hope our dataset can further boost the research in this direction.

 To summarize, our contributions are three-fold:

\begin{itemize}
\item We propose a positive point selection mechanism for segmenting blood vessels from angiographic X-ray images using SAM.
\item We introduce an algorithm for detecting stenoses and aneurysms over binary masks of the vessels.
\item We introduce a new benchmark dataset containing X-ray images of peripheral vessels along with the vessel binary masks and anomaly point labels.
\end{itemize}

%% file: sec/2_related_work.tex
\section{Related work}
\label{sec:relatedwork}

\subsection{Segmentation of blood vessels in x-ray images}

In our study, we explored both image processing \cite{Zong-Xian2022} and learning-based \cite{SAM} methods for angiography image segmentation. 
The lack of data in the community and also the novelity of the problem itself are limiting the fine-tuning or training specialized models.
Hence we choose to go with a zero-shot approach by leveraging pre-trained segmentation models. 
The Segment Anything Model (SAM) \cite{SAM} from Meta AI showcases the advancement, providing a system that can identify and segment a wide range of objects without needing prior training on them.

Similar to this paper, some existing works \cite{Chengliang2023, MedSAM, cheng2023sammed2d, wang2023sammed3d} leverage SAM for medical image segmentation. 
However they majorly choose to fine-tune SAM on medical 2D and 3D images of wide range, including ophtalmology images.
\cite{Guoyao2023} employs multi-box prompts to segment the optic disc.
In \cite{Haojian2023} the authors use SAM to annotate their dataset for training a new network for OCTA vessel segmentation.
\cite{Zhongxi2023} suggests a new learnable prompt layer for segmenting ophthalmology images.
In \cite{Junde2023} the authors have trained a model on 64 open-source medical datasets and added prompt options.
 
Regardless impressive results of the previous works, most of them either segment convex regions in medical images or vessels of different regions.
To the best of our knowledge Dr-SAM is the first end-to-end pipeline for angiography image analysis, including the vessel segmentation stage specified on \textit{peripheral vascular angiography} images.
Our contribution on segmentation part extends SAM's utility through a novel methodology of positive point selection which, along with user-specified bounding boxes, is guiding SAM for refined vascular segmentation.

\subsection{Anomaly detection}

Some previous works \cite{Iyer2021,Hampe2022,Moon2021} also touch the anomaly detection problem on medical images.
To detect anomalies, AngioNet \cite{Iyer2021} segments vessels and calculates the minimum and maximum diameters within each segment. 
In contrast, our approach identifies extremum points within each segment and designates anomalies in the areas between these extremum points. 
For finding diameters, we use skeleton detection algorithm \cite{topoligicalskelton}, which is a tool used for thinning or skeletonizing objects within an image to a single-pixel wide skeleton. 
The algorithm iteratively removes pixels from the edges of objects until only the minimal set of pixels that constitutes the "skeleton" remains, preserving the topology and general shape of the original object.

In \cite{Hampe2022} the authors utilize Coronary CT Angiography (CCTA) to extract coronary artery characteristics and assess stenosis significance with a CNN, focusing on artery geometry's impact on blood flow and local appearance for accurate stenosis assessment. 
In \cite{Moon2021} the authors enhance key frame detection with vessel extraction and employ CNN models with self-attention modules to classify stenosis, validating the algorithm through extensive cross-validation and external dataset evaluation, highlighting the use of heatmaps for visualization.
These methods illustrate the evolving complexity and specificity of techniques in detecting coronary anomalies, contrasting with our extremum point identification strategy for anomaly detection.

%% file: sec/3_method.tex
\section{Method}
In this section we introduce Dr-SAM, a universal algorithm for anomaly detection in blood vessels, incorporating zero-shot technology for vessel extraction followed by anomaly detection with the integration of topological skeleton.
Moreover, here we also present our benchmark dataset collected for thorough evaluation of our method and other approaches.

\begin{figure*}[htb]
    \centering\
    \includegraphics[width=\linewidth]{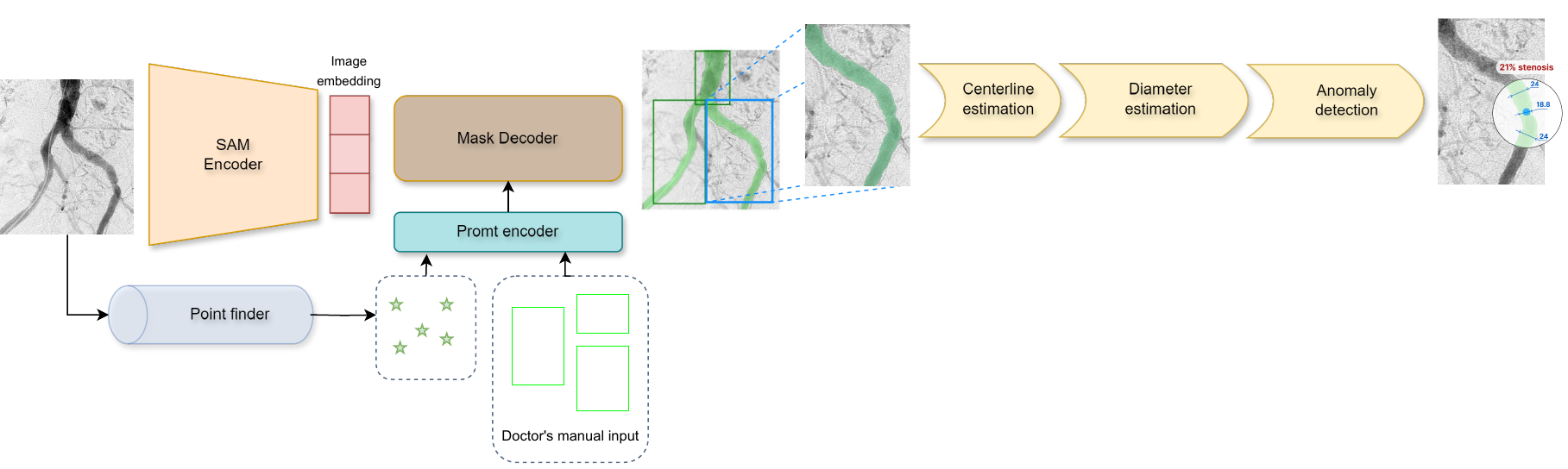}
    \caption{The overview of \textbf{\textit{Dr-SAM}}: For each bounding box provided by the user, first we determine \textit{five} positive points using our \textit{point finder} algorithm. This is followed by \textit{vessel extraction} by SAM \cite{SAM} conditioned on our positive points.
    Then for \textit{anomaly detection} we extract the centerline of the binary mask, obtained from the previous stage, by finding its skeleton, and use that skeleton for estimating vessel diameters, which are later being used to detect anomaly points on the vessels.}
    \label{fig:method}
\end{figure*}

Our streamlined approach for anomaly detection not only reduces computational costs having pipeline without training process, but also ensures applicability demonstrating effectiveness on angiographic images.
The overview of the framework can be found in Fig. \ref{fig:method}.

\subsection{Vessel extraction using point-conditioned SAM} 
We employ SAM \cite{SAM} for extracting vessels from X-ray images. 
In order to enhance results, we tailor SAM's prompt, particularly focusing on \textit{input points}. 
In our approach, we propose a novel algorithm for identifying positive points to feed SAM during segmentation (see Fig. \ref{fig:method}).

It is worth to mention that due to our experiments, incorporating negative points does not significantly impact segmentation outcomes hence we choose to design a special algorithm only for positive point selection. 
For the initial point, we select the most probable vessel point within the bounding box  through the following procedure:
\begin{enumerate}
    \item We generated a probability map for each pixel being a vessel pixel, assigning higher probabilities to pixels with lower values by assuming that darker points in the X-ray angiographic images are part of the vessel. We obtained the probability map by scaling and reverting the input values to [0, 1] (255 to 0 and 0 to 1, i.e. $x\mapsto 1-x/255 = \mbox{Probability}(x \mbox {is a vessel pixel})$ for $x\in [0,255]$).
    \item Excluded the pixels with probabilities lower than pre-defined threshold.
    \item Sampled 100 random points from the remaining set to prevent concentration of points in densely populated areas.
    \item Selected a point with the highest number of neighboring points within the predefined radius of $SelectionRadius$ from the sampled set as a positive point.
\end{enumerate}
For the second step, we select the most probable point that lies outside the region defined by the $ExcludeRadius$ from the first point and with the radius of $SecondPointSelectionRadius$.

To further enhance results, we implement a repeatable algorithm for identifying positive points. In simple words, the algorithm starts segmenting images by available points described in the previous paragraph. After segmentation, for selecting the next positive point, the algorithm avoids considering points on the previously predicted mask, selecting  the most possible positive point from the remaining image in the same way as described in the previous paragraph. For each iteration algorithm uses all available positive points collected from previous steps for segmentation, thereby ensuring consistent results after each iteration. However, to preserve previously segmented good results with minimal changes, we repeat this process three times at the result having overall 5 positive points, including 2 points from last paragraph, for the final segmentation.

\subsection{Anomaly detection using topological skeleton}
For anomaly detection, we utilized a topological skeleton \cite{topoligicalskelton}, a method widely employed in X-ray image studies within Computer Vision. The topological skeleton is a vital component in identifying or approximating the centerline of a vessel, aiding in the determination of its diameter in specific regions. 
Our proposed algorithm involves the use of a topological skeleton, which is subsequently pruned by removing unnecessary branches while preserving the vessel's structural integrity. 

To do so, our algorithm uses PlantCV \cite{gehan2017plantcv} techniques to extract branches from the topological skeleton. After getting the branches separated, we identified low-length branches. Our consideration was that branches with the length less then $MinBranchLength$ are not real branches of the blood vessel and were generated because of the anomalies in it. Removing low-length branches, we get tree-like structured skeleton of the blood vessel, which is better to estimate diameters along the segments. 

Further, we leveraged extracted segments of the skeleton to examine anomaly regions within the vessel segment. 
Treating segment-approximated diameters as values of a function, we conducted anomaly detection on each segment. 
Our primary consideration was that anomaly points constitute a subgroup of extremum points of the real function. 
However, the challenge arises from noise, which generates inaccuracies in the sequence's highs and lows. The primary contribution of our approach lies in mitigating noise by clustering close values in one region, facilitating the more accurate identification of highs and lows. 
By applying a threshold of $MinChangeThreshold$ to the variations between these extremum point values and the mean of surrounding point values, we identify anomaly points within the segment. 
For details see Algorithm \ref{alg:anomalydetection}.

After detecting the anomaly points we leverage the distance transform \cite{rosenfeld1969distance} technique to better estimate the diameters near the anomaly points. 
To do so for each centerline point we calculate the distance between that point and the nearest non-vessel point in the image by so estimating the radius of the vessel at that particular location of centerline.
After that we estimate the percentage of the anomaly as follows: 
\begin{enumerate}
        \item Extract branch from the skeleton as an array of points.
        \item Calculate the distance transform of anomaly point. 
            \begin{equation}
                dt_p = distanceTransform(segment[i])
            \end{equation}
        \item Calculate the distance transform of points located before and after anomaly point with the step equal to $length(segment)//5$. 
            \begin{equation}
                dt_{e1} = distanceTrasform(segment[i-step] 
            \end{equation}
            \begin{equation}
                dt_{e2} = distanceTransform(segment[i+step])
            \end{equation}
        \item For getting the percentage of the anomaly:
            \begin{equation}
                change_p = \frac{abs(mean(dt_{e1}, dt_{e2}) - dt_p)}{mean(dt_{e1}, dt_{e2})} 
            \end{equation}
\end{enumerate}
And finally, to get the type of the anomaly, we compare $mean(dt_{e1}, dt_{e2})$ and $dt_p$. If the first is more than the second, we have stenosis and our percentage change will be with $-$ sign. Otherwise, we will have an aneurysm, and the percentage change will have $+$ sign.

\begin{algorithm}
    \scriptsize
    \SetKwFunction{isOddNumber}{isOddNumber}
    \SetKwInOut{KwIn}{Input}
    \SetKwInOut{KwOut}{Output}

    \KwIn{2D Predicted binary mask}
    \KwOut{List of anomaly points}

    $anomalyPoints = [\ ]$\;
    $segmentedSkeleton \leftarrow segmentize(skeletonize(mask))$\;
    $skeletonThickness \leftarrow getThickness(mask)$\;
    
    \For{$seg$ \textbf{in} $segmentedSkeleton$}{
        $seg \leftarrow segmentedSkeleton[i]$\;
        $thickness \leftarrow skeletonThickness[seg]$\;
        $eps \leftarrow length(seg)/10$\;
        $minSamples \leftarrow 1$\;
        $localExtremums \leftarrow findLocalExtremums(thickness)$\;
        $clusters \leftarrow DBSCAN(localExtremums, eps, minSamples)$\;
        
        \BlankLine
        
        $newExtremums = [\ ]$\;
        \For{$c$ \textbf{in} $clusters$}{
            $indices \leftarrow c.indices$\;
            $center \leftarrow int(mean(indices))$\;
            $newExtremums.add([center,thickness[center]])$
        }
        
        \BlankLine

        $filteredExtremums = [\ ]$\;
        $N \leftarrow length(newExtremums)$\;
        \For{$i \gets 1$ \KwTo $N-1$}{
            $current \leftarrow filteredExtremums[i][1]$\;
            $previous \leftarrow filteredExtremums[i-1][1]$\;
            $next \leftarrow filteredExtremums[i+1][1]$\;
            \If{$current < previous \And current < next$}{
                $filteredExtremums.add(current)$\;
            }
            \If{$current > previous \And current > next$}{
                $filteredExtremums.add(current)$\;
            }
        }

        \BlankLine

        $radius \leftarrow length(seg)//5$\;
        \For{$point$ \textbf{in} $filteredExtremums$}{
            $index \leftarrow point[0]$\;
            $meanThickness \leftarrow (thickness[index-radius] + thickness[index+radius])/2$\;
            
            \If{$abs(point[1] - meanThickness)/meanThickness > 0.5$}{
                $anomalyPoints.add(seg[index])$\;
            }
        }
    }
    \KwRet{$anomalyPoints$}
    \caption{Anomaly detection algorithm}
    \label{alg:anomalydetection}
\end{algorithm}

\subsection{Benchmark Dataset}

Our dataset consists of carefully selected images from 500 angiographic examinations of the pelvic-iliac arteries, carried out between 2018 and 2024 at Bad Oeynhausen Hospital and JWK Klinikum Minden, within their radiology departments. The focus of these examinations was the abdominal aorta below the renal arteries and the pelvic arteries.
Using Adobe Lightroom, only the pertinent areas of these examinations were cropped to isolate the regions of interest. Of these images, 450 have a resolution of 386x448 pixels, and 50 have a resolution of 819x950 pixels. The dataset includes 170 images featuring at least one stenosis and 64 images with at least one aneurism.
Following this initial selection, Adobe Photoshop \cite{adobephotoshop} was employed to create a vessel mask for each cropped image, which outlines the arterial structure. 
Additionally, any narrowing and widening observed within the arterial regions were meticulously marked. This dataset is a comprehensive compilation that provides a significant resource for studying the conditions affecting the pelvic-iliac arteries, demonstrating a targeted approach to vascular imaging research.

%% file: sec/4_experiments.tex
\section{Experiments}
In this section we first discuss some implementation details of Dr-SAM, then make thorough analysis of its segmentation and anomaly detection (including the centerline detection, diameter and anomaly estimations) stages.

\begin{figure*}
    \centering\
    \includegraphics[width=\linewidth,height=21cm,keepaspectratio]{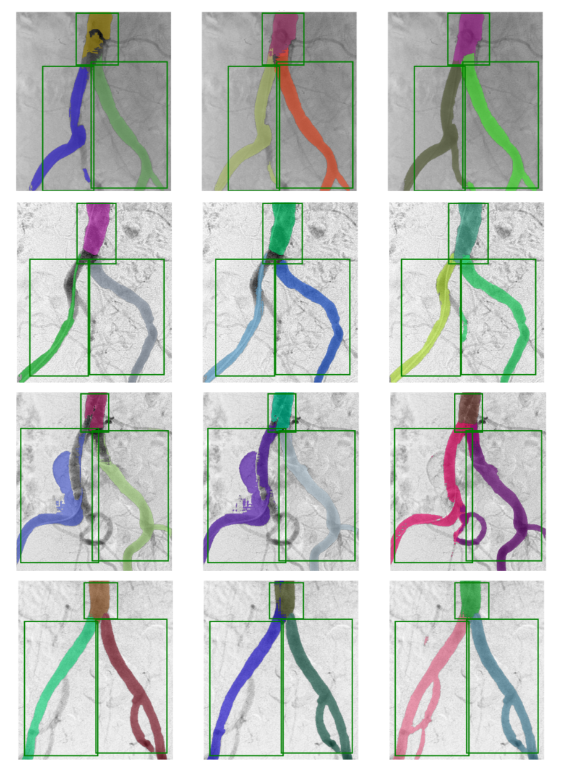}
    \caption{Qualitative comparison of three different approaches for segmentation. From left to right: SAM, naive approach of selecting the positive point as the most probable vascular pixel, our method}
    \label{fig:comparison}
\end{figure*}

\begin{figure*}[htb]
    \centering\
    \includegraphics[width=\linewidth,height=21cm,keepaspectratio]{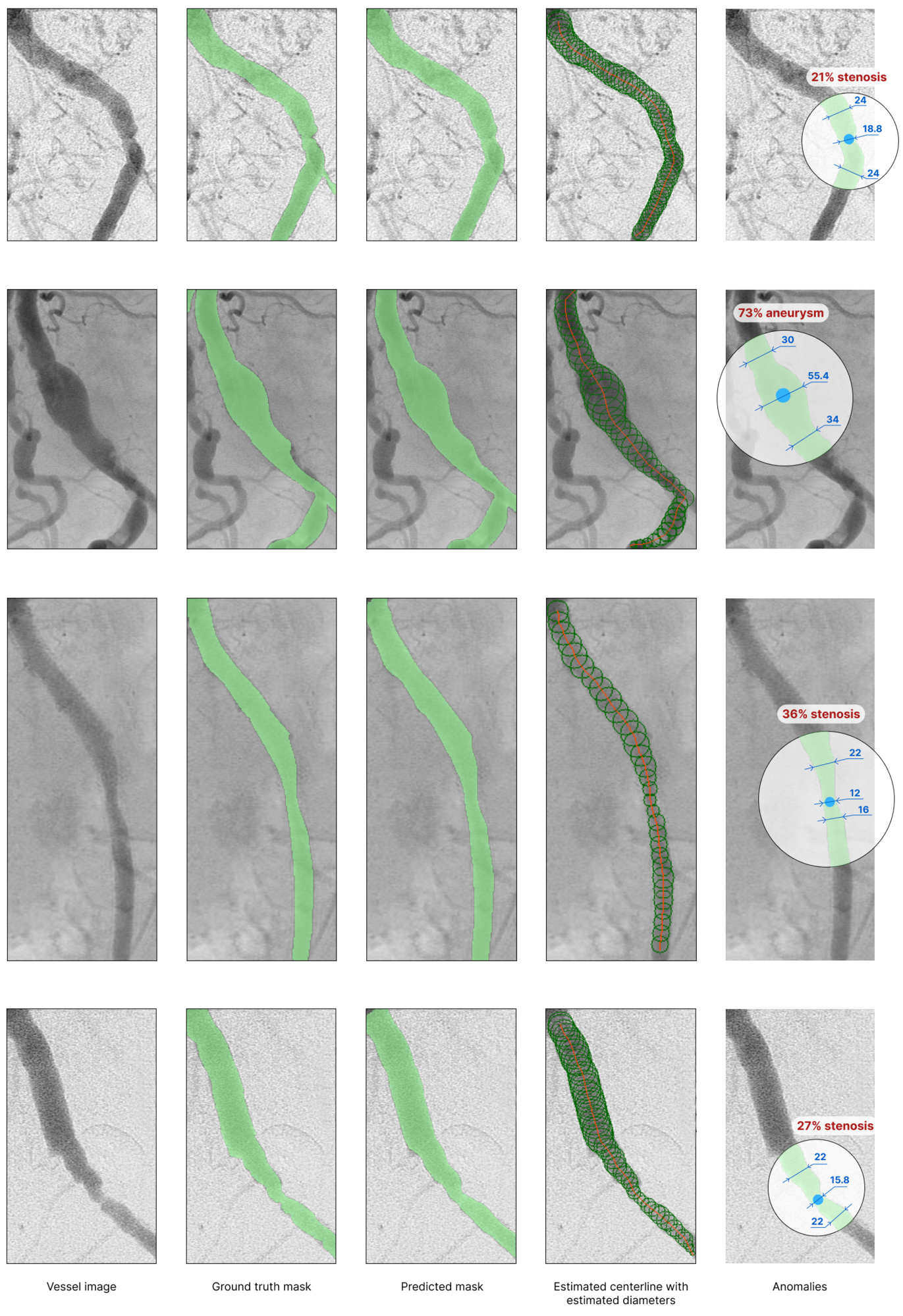}
    \caption{Dr. SAM pipeline results for each step}
    \label{fig:method_results}
\end{figure*}

\subsection{Implementation details}

The experiments were conducted on a diverse set of angiography images, encompassing various structures and anomalies. We utilized angiographic X-ray images with bounding boxes and anomaly points validated by two professional doctors in the field of vascular imaging. Throughout the experiments, we configured parameters as follows: $SelectionRadius=75$, $SecondPointSelectionRadius = 50$, $ExcludeRadius = 100$, $MinBranchLength = 40$, $MinChangeThreshold = 50\%$. Our code is implemented using PyTorch. The mean time for the segmentation part takes $0.66$ seconds on average, while the anomaly detection part takes $0.65$ seconds on average.

\subsection{Segmentation Analysis}

For the segmentation aspect, we aim to evaluate our methodology against two established techniques:
The first technique involves applying the Segment Anything Model (SAM) to the original images, employing bounding boxes as prompts without any  enhancement or additional prompts.
The second, a naive positive point selection approach, attempts to identify the pixel with the lowest value (i.e. the highest probability of being a vascular pixel) within each specified bounding box on the original image (normalized to have values from the range [0,1]), and gives it as a positive point prompt.
We present the analysis both visually and through quantitative measures. 
For our quantitative metric, we use the Intersection over Union (IoU) \cite{Jaccard1901IOU} to compare our predictions with the ground truths.
The ground truth is derived from binary masks of each box, meticulously annotated by experts. 

In Tab. \ref{table:quantitative_analysis} we present the mean Intersection over Union calculated on \textbf{\textit{450 various angiography images}} segmented by 1. standalone SAM; 2. SAM with additional positive point naively chosen as the highest probable point of being a vascular pixel; 3. SAM with additional positive points gathered by our method.
The experiments clearly show the advantage of our method for segmenting the vessels in angiography images.

\begin{table}[ht]
\centering
\begin{tabular}{|c|c|c|c|}
\hline
         & \textbf{SAM} & \textbf{Naive approach} & \textbf{Our method} \\ \hline
\textbf{MIoU}   & 0.754   &    0.807   & \textbf{0.859}   \\ \hline
\end{tabular}
\caption{Quantitative comparison by using the mean IoU metric.}
\label{table:quantitative_analysis}
\end{table}

We also perform qualitative analysis of our method by comparing it with the above mentioned approaches of vanilla SAM and highest probable positive point selection approach. 
Fig. \ref{fig:comparison} shows the clear advantage of our method in comparison, different colors are applied for more visually appealing demonstration.

\subsection{Anomaly detection}
\begin{figure}
  \centering
   \includegraphics[width=\linewidth]{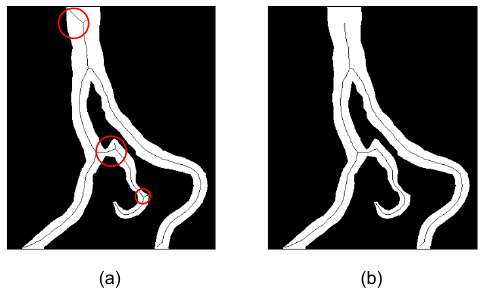}

   \caption{Topological skeleton in (a), and processed skeleton by our algorithm (b)}
   \label{fig:skeleton}
\end{figure}

For centerline estimation, our objective is to assess its performance relative to the topological skeleton. As depicted in Figure \ref{fig:skeleton}, the distinction between skeletons is clearly illustrated. Our proposed algorithm effectively eliminates extra branches from the skeleton (indicated by red circles) that do not belong to the vessel structure.

In Figure \ref{fig:method_results}, we present the successful centerline estimation results. These results clearly demonstrate the adaptability of the centerline algorithm to anomalies, leading to improved accuracy in diameter estimation. 
Following successful centerline detection, the distance transform method yields excellent results in diameter estimation, including in challenging anomal regions. 
And finally, our anomaly detection algorithm excellently finds both stenosis and aneurysm parts of the segments, giving good results in indicating the level of anomaly.

%% file: sec/5_conclusion.tex
\section{Conclusion}
In this paper, we present Dr-SAM, a novel approach for an end-to-end detection of stenoses and aneurisms in peripheral blood vessels. 
We introduce a customized positive point detection for Segment Anything Model (SAM) to capture blood vessels in X-ray angiographic images without the need for any additional training.

Through a series of experiments, we demonstrate the effectiveness of our approach in vessel extraction and anomaly detection.
Our method offers significant advancements over existing segmentation techniques in this particular task.

Additionally, we introduce a benchmark dataset of 450 angiography images of peripheral vessels, annotated and labeled by highly qualified experts. 
We plan to make the dataset and the codes publicly available.

In conclusion, our work contributes to expanding research on medical image processing by introducing new ideas and tools for future work.
Addressing limitations of the naive approaches, our approach has the potential to advance the state-of-the-art in anomaly detection in blood vessels research, providing a more effective and efficient solution for a wide range of applications.

%% file: main.bbl
\begin{thebibliography}{23}
\providecommand{\natexlab}[1]{#1}
\providecommand{\url}[1]{\texttt{#1}}
\expandafter\ifx\csname urlstyle\endcsname\relax
  \providecommand{\doi}[1]{doi: #1}\else
  \providecommand{\doi}{doi: \begingroup \urlstyle{rm}\Url}\fi

\bibitem[{Adobe Systems Incorporated}(Year of Version)]{adobephotoshop}
{Adobe Systems Incorporated}.
\newblock \emph{Adobe Photoshop}, Year of Version.
\newblock Software available from Adobe Systems Incorporated.

\bibitem[Blum(1967)]{topoligicalskelton}
Harry Blum.
\newblock A transformation for extracting new descriptors of shape.
\newblock In \emph{Models for the Perception of Speech and Visual Form}, pages 362--380. MIT Press, Cambridge, Massachusetts, 1967.

\bibitem[Chen et~al.(2018)Chen, Zhu, Papandreou, Schroff, and Adam]{deeplabv3plus2018}
Liang-Chieh Chen, Yukun Zhu, George Papandreou, Florian Schroff, and Hartwig Adam.
\newblock Encoder-decoder with atrous separable convolution for semantic image segmentation.
\newblock In \emph{ECCV}, 2018.

\bibitem[Cheng et~al.(2023)Cheng, Ye, Deng, Chen, Li, Wang, Su, Huang, Chen, Sun, He, Zhang, Zhu, and Qiao]{cheng2023sammed2d}
Junlong Cheng, Jin Ye, Zhongying Deng, Jianpin Chen, Tianbin Li, Haoyu Wang, Yanzhou Su, Ziyan Huang, Jilong Chen, Lei Jiangand~Hui Sun, Junjun He, Shaoting Zhang, Min Zhu, and Yu Qiao.
\newblock Sam-med2d, 2023.

\bibitem[Deng et~al.(2023)Deng, Zou, Ren, Wang, Yuan, Ying, and Fu]{Guoyao2023}
Guoyao Deng, Ke Zou, Kai Ren, Meng Wang, Xuedong Yuan, Sancong Ying, and Huazhu Fu.
\newblock Sam-u: Multi-box prompts triggered uncertainty estimation for reliable sam in medical image.
\newblock \emph{arXiv:2307.04973}, 1, 2023.

\bibitem[Franchi et~al.(2008)Franchi, Gallo, and Placidi]{digitalprocessing}
Danilo Franchi, Pasquale Gallo, and Giuseppe Placidi.
\newblock A novel segmentation algorithm for digital subtraction angiography images: First experimental results.
\newblock In \emph{Advances in Visual Computing}, pages 612--623, Berlin, Heidelberg, 2008. Springer Berlin Heidelberg.

\bibitem[Gehan et~al.(2017)Gehan, Fahlgren, Abbasi, Berry, Callen, Chavez, Doust, Feldman, Gilbert, Hodge, Hoyer, Lin, Liu, Liz{\'a}rraga, Lorence, Miller, Platon, Tessman, and Sax]{gehan2017plantcv}
Malia~A Gehan, Noah Fahlgren, Arash Abbasi, Jeffrey~C Berry, Steven~T Callen, Leonardo Chavez, Andrew~N Doust, Max~J Feldman, Kerrigan~B Gilbert, John~G Hodge, J~Steen Hoyer, Aanika Lin, Siobhan Liu, Carolina Liz{\'a}rraga, Argelia Lorence, Michael Miller, Eric Platon, Monica Tessman, and Tony Sax.
\newblock Plantcv v2: Image analysis software for high-throughput plant phenotyping.
\newblock \emph{PeerJ}, 5:\penalty0 e4088, 2017.

\bibitem[Jaccard(1901)]{Jaccard1901IOU}
Paul Jaccard.
\newblock Distribution de la flore alpine dans le bassin des dranses et dans quelques régions voisines.
\newblock \emph{Bulletin de la Société Vaudoise des Sciences Naturelles}, 37:\penalty0 241--272, 1901.

\bibitem[K. et~al.(2021)K., P., A., J., R., Subban, and Figueroa]{Iyer2021}
Iyer K., Najarian~C. P., Fattah~A. A., Arthurs~C. J., Soroushmehr S.~M. R., V. Subban, and C.~A. Figueroa.
\newblock Angionet: a convolutional neural network for vessel segmentation in x-ray angiography.
\newblock \emph{Scientific Reports}, 11(1), 2021.

\bibitem[Kirillov et~al.(2023)Kirillov, Mintun, Ravi, Mao, Rolland, Gustafson, Xiao, Whitehead, Berg, Lo, Doll{\'a}r, and Girshick]{SAM}
Alexander Kirillov, Eric Mintun, Nikhila Ravi, Hanzi Mao, Chloe Rolland, Laura Gustafson, Tete Xiao, Spencer Whitehead, Alexander~C. Berg, Wan-Yen Lo, Piotr Doll{\'a}r, and Ross Girshick.
\newblock Segment anything.
\newblock \emph{arXiv:2304.02643}, 2023.

\bibitem[Luo and Sun(2023)]{luo2023digital}
Yanping Luo and Linggang Sun.
\newblock Digital subtraction angiography image segmentation based on multiscale hessian matrix applied to medical diagnosis and clinical nursing of coronary stenting patients.
\newblock \emph{Journal of Radiation Research and Applied Sciences}, 16\penalty0 (3):\penalty0 100603, 2023.

\bibitem[Ma et~al.(2024)Ma, He, Li, Han, You, and Wang]{MedSAM}
Jun Ma, Yuting He, Feifei Li, Lin Han, Chenyu You, and Bo Wang.
\newblock Segment anything in medical images.
\newblock \emph{Nature Communications}, 15:\penalty0 1--9, 2024.

\bibitem[Moon et~al.(2021)Moon, Lee, Cha, Chung, Lee, Cho, and Choi]{Moon2021}
Jong~Hak Moon, Da~Young Lee, Won~Chul Cha, Myung~Jin Chung, Kyu-Sung Lee, Baek~Hwan Cho, and Jin~Ho Choi.
\newblock Automatic stenosis recognition from coronary angiography using convolutional neural networks.
\newblock \emph{Computer Methods and Programs in Biomedicine}, 198:\penalty0 105819, 2021.

\bibitem[N et~al.(2022)N, van Velzen~SGM, RN, JPS, C, JP, M, T, and I.]{Hampe2022}
Hampe N, van Velzen~SGM, Planken RN, Henriques JPS, Collet C, Aben JP, Voskuil M, Leiner T, and Išgum I.
\newblock Deep learning-based detection of functionally significant stenosis in coronary ct angiography.
\newblock \emph{Front Cardiovasc Med.}, 9:964355, 2022.

\bibitem[Ning et~al.(2023)Ning, Wang, Chen, and Li]{Haojian2023}
Haojian Ning, Chengliang Wang, Xinrun Chen, and Shiying Li.
\newblock An accurate and efficient neural network for octa vessel segmentation and a new dataset.
\newblock \emph{arXiv:2309.09483}, 1, 2023.

\bibitem[Qiu et~al.(2023)Qiu, Hu†, Li, and Liu]{Zhongxi2023}
Zhongxi Qiu, Yan Hu†, Heng Li, and Jiang Liu.
\newblock Learnable ophthalmology sam.
\newblock \emph{arXiv:2304.13425}, 1, 2023.

\bibitem[Ronneberger et~al.(2015)Ronneberger, P.Fischer, and Brox]{Unet}
O. Ronneberger, P.Fischer, and T. Brox.
\newblock U-net: Convolutional networks for biomedical image segmentation.
\newblock In \emph{Medical Image Computing and Computer-Assisted Intervention (MICCAI)}, pages 234--241. Springer, 2015.
\newblock (available on arXiv:1505.04597 [cs.CV]).

\bibitem[Rosenfeld and Pfaltz(1969)]{rosenfeld1969distance}
Azriel Rosenfeld and John~L Pfaltz.
\newblock Distance functions on digital pictures.
\newblock \emph{Pattern Recognition}, 1\penalty0 (1):\penalty0 33--61, 1969.

\bibitem[Thiriet et~al.(2015)Thiriet, Delfour, and Garon]{Thiriet2015}
Marc Thiriet, Michel Delfour, and Andr{\'e} Garon.
\newblock \emph{Vascular Stenosis: An Introduction}, pages 781--868.
\newblock Springer Berlin Heidelberg, Berlin, Heidelberg, 2015.

\bibitem[Wang et~al.(2023{\natexlab{a}})Wang, Chen, Ning, and Li]{Chengliang2023}
Chengliang Wang, Xinrun Chen, Haojian Ning, and Shiying Li.
\newblock Sam-octa: A fine-tuning strategy for applying foundation model to octa image segmentation tasks.
\newblock \emph{arXiv:2309.11758}, 1, 2023{\natexlab{a}}.

\bibitem[Wang et~al.(2023{\natexlab{b}})Wang, Guo, Ye, Deng, Cheng, Li, Chen, Su, Huang, Shen, Fu, Zhang, He, and Qiao]{wang2023sammed3d}
Haoyu Wang, Sizheng Guo, Jin Ye, Zhongying Deng, Junlong Cheng, Tianbin Li, Jianpin Chen, Yanzhou Su, Ziyan Huang, Yiqing Shen, Bin Fu, Shaoting Zhang, Junjun He, and Yu Qiao.
\newblock Sam-med3d, 2023{\natexlab{b}}.

\bibitem[Wu et~al.(2023)Wu, Zhu, Liu, Jin, and Xu]{Junde2023}
Junde Wu, Jiayuan Zhu, Yuanpei Liu, Yueming Jin, and Min Xu.
\newblock One-prompt to segment all medical images.
\newblock \emph{arXiv:2305.10300}, 3, 2023.

\bibitem[Yin and Xu(2022)]{Zong-Xian2022}
Zong-Xian Yin and Hong-Ming Xu.
\newblock An unsupervised image segmentation algorithm for coronary angiography.
\newblock \emph{BioData Mining}, 15\penalty0 (27), 2022.

\end{thebibliography}
